\documentclass[twoside]{article}
\usepackage{amsfonts,amssymb,amsbsy,textcomp,marvosym,picins,amsmath,caption,threeparttable,amsthm,subfigure}
\usepackage{eurosym,mathrsfs,fancyhdr,CJK,multicol,graphics,indentfirst,color,bm,upgreek,booktabs,graphicx}
\usepackage{multirow}
\usepackage{diagbox}
\renewcommand{\thefootnote}{\fnsymbol{footnote}}
\def\footnoterule{\kern 1mm \hrule width 10cm \kern 2mm}

\allowdisplaybreaks
\catcode`@=11
\def\title#1{\vspace{3mm}\begin{flushleft}\vglue-.1cm\Large\bf\boldmath\protect\baselineskip=18pt plus.2pt minus.1pt #1
\end{flushleft}\vspace{1mm} }
\def\author#1{\begin{flushleft}\normalsize #1\end{flushleft}\vspace*{-4pt} \vspace{3mm}}
\def\address#1#2{\begin{flushleft}\vglue-.35cm${}^{#1}$\small\it #2\vglue-.35cm\end{flushleft}\vspace{-2mm}\par}

\catcode`@=11
\def\section{\@startsection{section}{1}{\z@}%
 {-3ex \@plus -.3ex \@minus -.2ex}%
 {2.2ex \@plus.2ex}%
{\normalfont\normalsize\protect\baselineskip=14.5pt plus.2pt minus.2pt\bfseries}}
\def\subsection{\@startsection{subsection}{2}{\z@}%
 {-3ex\@plus -.2ex \@minus -.2ex}%
 {2ex \@plus.2ex}%
{\normalfont\normalsize\protect\baselineskip=12.5pt plus.2pt minus.2pt\bfseries}}
\def\subsubsection{\@startsection{subsubsection}{3}{\z@}%
 {-2.2ex\@plus -.21ex \@minus -.2ex}%
 {1.4ex \@plus.2ex}
{\normalfont\normalsize\protect\baselineskip=12pt plus.2pt minus.2pt\sl}}

\begin{document}
\begin{CJK*}{GBK}{song}
\thispagestyle{empty}
\vspace*{-13mm}
\noindent {\small Zhu DH, Dai XY, Chen JJ. Pre-train and Learn: Preserving Global Information for Graph Neural Networks.
JOURNAL OF COMPUTER SCIENCE AND TECHNOLOGY \ 36(6): 
\ 1420-1430 Nov. 2021. DOI 10.1007/s11390-020-0142-x}
\vspace*{2mm}

\title{Pre-train and Learn: Preserving Global Information for Graph Neural Networks}

\author{Dan-Hao Zhu$^{1,2}$, Xin-Yu Dai$^{2,*}$, and Jia-Jun Chen$^{2}$}
\address{1}{Library, Jiangsu Police Institute, Nanjing 210031, Jiangsu, China}
\address{2}{Department of Computer Science and Technology, Nanjing University, Nanjing 210093, Jiangsu China}

\vspace{2mm}

\noindent E-mail: zhudanhao@jspi.cn; {daixinyu,chenjj}@nju.edu.cn \\[-1mm]

\noindent Received October 10, 2019 ; revised May 12, 2020.\\[1mm]

\let\thefootnote\relax\footnotetext{{}\\[-4mm]\indent\ Regular Paper\\[.5mm]
\indent\ This work was partially supported by the Natural Science Foundation of the Jiangsu Higher Education Institutions of China under grant No 18kJB510010, Social Science Foundation of Jiangsu Province under grant No 19TQD002, and National Nature Science Foundation of China (NSFC) under grant number 61976114. \\[.5mm]
\indent\ $^*$Corresponding Author
\\[.5mm]\indent\ \copyright Institute of Computing Technology, Chinese Academy of Sciences \& Springer Nature Singapore Pte Ltd. 2020}

\noindent {\small\bf Abstract} \quad  {\small \textcolor{blue}{Graph neural networks (GNNs) have shown great power in learning on graphs. However, it is still a challenge for GNNs to model information faraway from the source node. The ability to preserve global information can enhance graph representation and hence improve classification precision. In the paper, we propose a new learning framework named G-GNN (Global information for GNN) to address the challenge. First, the global structure and global attribute features of each node are obtained via unsupervised pre-training, which preserve the global information associated to the node. Then, using the pre-trained global features and the raw attributes of the graph, a set of parallel kernel GNNs is used to learn different aspects from these heterogeneous features. Any general GNN can be used as a kernal and easily obtain the ability of preserving global information, without having to alter their own algorithms. Extensive experiments have shown that state-of-the-art models, e.g. GCN, GAT, Graphsage and APPNP, can achieve improvement with G-GNN on three standard evaluation datasets. Specially, we establish new benchmark precision records on Cora (84.31\%) and Pubmed (80.95\%) when learning on attributed graphs.}}

\vspace*{3mm}

\noindent{\small\bf Keywords} \quad {\small Graph neural network, Network embedding, Representation learning }

\vspace*{4mm}

\end{CJK*}
\baselineskip=18pt plus.2pt minus.2pt
\parskip=0pt plus.2pt minus0.2pt
\begin{multicols}{2}

\section{Introduction}
Semi-supervised learning on graphs is popular in various real world applications, since labels are often expensive and difficult to collect. In the recent years, Graph neural networks (GNNs) have shown great power in the semi-supervised learning on attributed graphs. GNN often contains multiple layers, and the nodes collect information from the neighborhood iteratively, layer by layer. The representative methods include graph convolutional network (GCN)~\cite{kipf2016semi}, Graphsage~\cite{hamilton2017inductive}, graph attention networks~\cite{velivckovic2017graph} and so on.

Due to the multi-layer message aggregation scheme, GNN is easy to be over-smoothing after a few propagation steps~\cite{li2018deeper}. That is to say, the node representations tend to be identical and lack of distinction. In general, GNNs can afford only 2 layers to collect information within 2 hops of neighborhood, otherwise the over-smoothing problem will deteriorate the performance.  Several previous works~\cite{abu2018n,klicpera2018predict} aimed to address the problem by expanding the size of utilized neighborhood. For example, N-GCN~\cite{abu2018n} trains multiple instances of GCNs over node pairs discovered at different distances in random walks. PPNP/APPNP~\cite{klicpera2018predict} introduces the teleport probability of personalized PageRank.

These solutions~\cite{abu2018n,klicpera2018predict} are purely based on semi-supervised learning, which have their natural bottleneck in modeling global information. As the reception field is getting larger and more nodes are involved, more powerful models are required to explore the complex relationships among nodes. However, the labels for training are quite sparse in the semi-supervised learning, and hence cannot afford to train models with high complexity well. Hence, the existing works often have to restrain the complexity of models, which limits their ability in learning global information.

The unsupervised learning methods based on random walk, e.g. Deepwalk~\cite{perozzi2014deepwalk} and Node2vec~\cite{grover2016node2vec}, can be used to obtain the global structure information of each node. These methods first sample node sequences that contain the structure regularity of the network, then try to maximize the likelihood of neighborhood node pairs within certain distance, e.g. 10 hops. Hence, they can capture global structure features without label information.

In this paper, we propose a new learning schema of \emph{pre-training and learning} to address the global information preserving problem in semi-supervised learning. First, instead of improving the aggregation function via semi-supervision, we obtain the global structure and global attribute features by pre-training the graph with random walk strategy in the unsupervised learning. Second, we design a GNN based framework to conduct semi-supervised classification by learning from the pre-trained features and the original graph.

The two-stage schema of  pretraining-and-learning has several advantages. First, the global information modeling procedure is decoupled with the subsequent semi-supervised learning method. Therefore, the modeling of the global information no longer suffers from the sparse supervision or over-smoothing problem. Moreover, a general GNN can enhance its global information preserving ability by applying our learning framework, without having to altering its algorithm. Second, the proposed framework takes the advantage of both random walk and GNN, which can not only utilize global information but also aggregate local information well. Moreover, the framework allows GNN to be applied to plain graphs without attributes, since the unsupervised structure features can be used as graph attributes directly.

In all, our contributions are as follows.
\begin{itemize}
    \item We propose a learning framework named Global information for Graph Neural Network (G-GNN) for semi-supervised classification on graphs. The proposed pretraining-and-learning schema allows GNN models to use global information for learning, without altering their algorithms. Moreover, the schema enables GNN to be applied to plain graphs.

    \item We design the global information as the global structure and global attribute features to each node, and propose a set of parallel GNNs to learn different aspects from the pretrained global features and the original graph.

    \item Our method achieves state-of-the-art results in semi-supervised learning on both plain and attributed graphs. Specially, the precisions of attributed graph learning on Cora (84.31\%) and Pubmed (80.95\%) are the new benchmark results.
\end{itemize}

The rest of the paper is organized as follows. In section 2, the preliminaries are given. We introduce our method in section 3. Section 4 presents the experiments. Section 5 briefly summarizes related work. Finally, we conclude our work in section 6.

\section{Preliminary}
\subsection{Definition}
First, we will give the formal definition of attributed/plain graph, and the problem we are going to solve.

Let $G=(V,E)$ be a graph, where $V=\{v_1,v_2...v_n\}$ denotes the node set, $E$ denotes the edges with the adjacency matrix $\boldsymbol A \in \mathbb{R}^{n \times n}$. If $ A_{ij}$ is not equal to 0, there is a link from $v_i$ to $v_j$ with weight $A_{ij}$. If $G$ is an attributed graph, there is a corresponding attribute matrix $\boldsymbol X \in \mathbb{R}^{n \times f}$, where the $i$th row denotes $v_i$'s attributes and $f$ denotes the total amount of attributes. If $G$ is a plain network, no $\boldsymbol X$ is provided. The graph contains label information $\boldsymbol Y \in \mathbb{R}^{n  \times c}$, where the $i$th row denotes $v_i$'s one hot label vector.  The amount of labels is $c$.

During the training stage, the provided data is the entire adjacency matrix $\boldsymbol A$ and the node attributes $\boldsymbol X$. Only labels of the training nodes $V_{train} \subset V$ are given. The task of the semi-supervised learning in attributed graph is to predict the rest of the node labels $\boldsymbol Y_{V_{\rm   train}}$. $\boldsymbol X$ is not provided for plain graph learning.

\subsection{Graph Neural Networks}
We will introduce how general GNNs solve the semi-supervised learning problem. Note that current GNNs can only be applied to attributed graphs. Therefore, we assume $\boldsymbol X$ is given here.

Among the huge family of GNNs, Graph convolutional network (GCN)~\cite{kipf2016semi} is a simple and pioneering method. Let $\boldsymbol{\hat{A}} = \boldsymbol A + \boldsymbol I_n$ be the adjacency matrix with self-connections, where $\boldsymbol I_n$ is an identity matrix. The self-loops will allow GCN to consider attributes of the represented nodes when aggregating the neigbhood's attributes. Let $\boldsymbol{\hat{\hat{A}}} = \boldsymbol{\hat{D}}^{-1/2}\boldsymbol{\hat{A}} \boldsymbol{\hat{D}}^{-1/2}$ be the nomalized adjacency matrix, where $\boldsymbol{\hat{D}}$ denotes the diagonal degree matrix where ${\hat{D}}_{ii} = \sum_j {\hat{A}}_{ij}$. The two-layer GCN produces hidden states by aggregating neighborhood attributes iteratively, as in (1).
\begin{equation}
    \boldsymbol H_{{\rm GCN}} = \boldsymbol{\hat{\hat{A}}}{ Relu}(\boldsymbol{\hat{\hat{A}}XW_0})\boldsymbol W_1 .
\end{equation}
where $\boldsymbol H_{\rm GCN} \in \mathbb{R}^{n  \times c}$ and $Relu(.)$ is an activation function commonly used in neural networks. Each row in $\boldsymbol H_{\rm GCN}$ denotes the final hidden states of a node, and each row corresponds to a prediction catagory. $\boldsymbol{W_0}$ and $\boldsymbol{W_1}$ are the trainable weight matrices. After that, the classification probability on each class $\boldsymbol Z_{{\rm GCN}}$ is obtained via $softmax(.)$, a normalization function commonly used in machine learning,  as in (2).
\begin{equation}
\boldsymbol Z_{{\rm GCN}}=softmax(\boldsymbol H_{{\rm GCN}} ).
\end{equation}
Finally, a loss function is applied to measure the difference between the predict probability and the ground truth labels.

Many of the following studies aim to improve the aggregation function, such as assigning alternative weights to the neighborhood nodes~\cite{velivckovic2017graph}, adding skip connections~\cite{hamilton2017inductive}, introducing teleport probability~\cite{klicpera2018predict} and so on. These methods can be viewed as a transform from the original $\boldsymbol{X}$ and $\boldsymbol{A}$ to the final hidden states $\boldsymbol{H}$, as in (3).

\begin{equation}
{GNN}(\boldsymbol{X},\boldsymbol{A}): \boldsymbol{X},\boldsymbol{A} \rightarrow \boldsymbol{H}.
\end{equation}

From (1), we can see only 2-hops of local information can be used. The over-smoothing problem prevents from adding more layers, so the global information is difficult to be integrated in.
The input attributes $\boldsymbol{X}$ are necessary in the general learning framework of GNN. Hence, GNN cannot be applied to plain graphs directly. In the next section, we will show how to solve these problems with the proposed pretraining-and-learning schema.

\section{The Proposed Framework of G-GNN}
In the section, we first give an overview of G-GNN within the context of attributed graph. Second, the method to obtain the global features is introduced. Third, a parallel GNN based method is proposed to learn from all these features. Finally, we show how to extend G-GNN to plain network.
\subsection{Overview}
The overview of the G-GNN is shown in Fig. 1. First, the global structure feature matrix $\boldsymbol X^{(s)}$ and attribute feature matrix $\boldsymbol X^{(a)}$ are learned in an unsupervised way. Next, $\boldsymbol X^{(s)}$, $\boldsymbol X^{(a)}$ and the original attribute matrix $\boldsymbol X$ are fed to a parallel GNN based model, to learn their corresponding hidden states. Finally, the final hidden states are the weighted sum of the 3 hidden states $\boldsymbol H^{(s)}$, $\boldsymbol X^{(a)}$ and $\boldsymbol H^{(o)}$.

\begin{figure*}[tb]
\label{fig:1}
\begin{center}
{\includegraphics[width=0.8\linewidth,height=6.7cm]{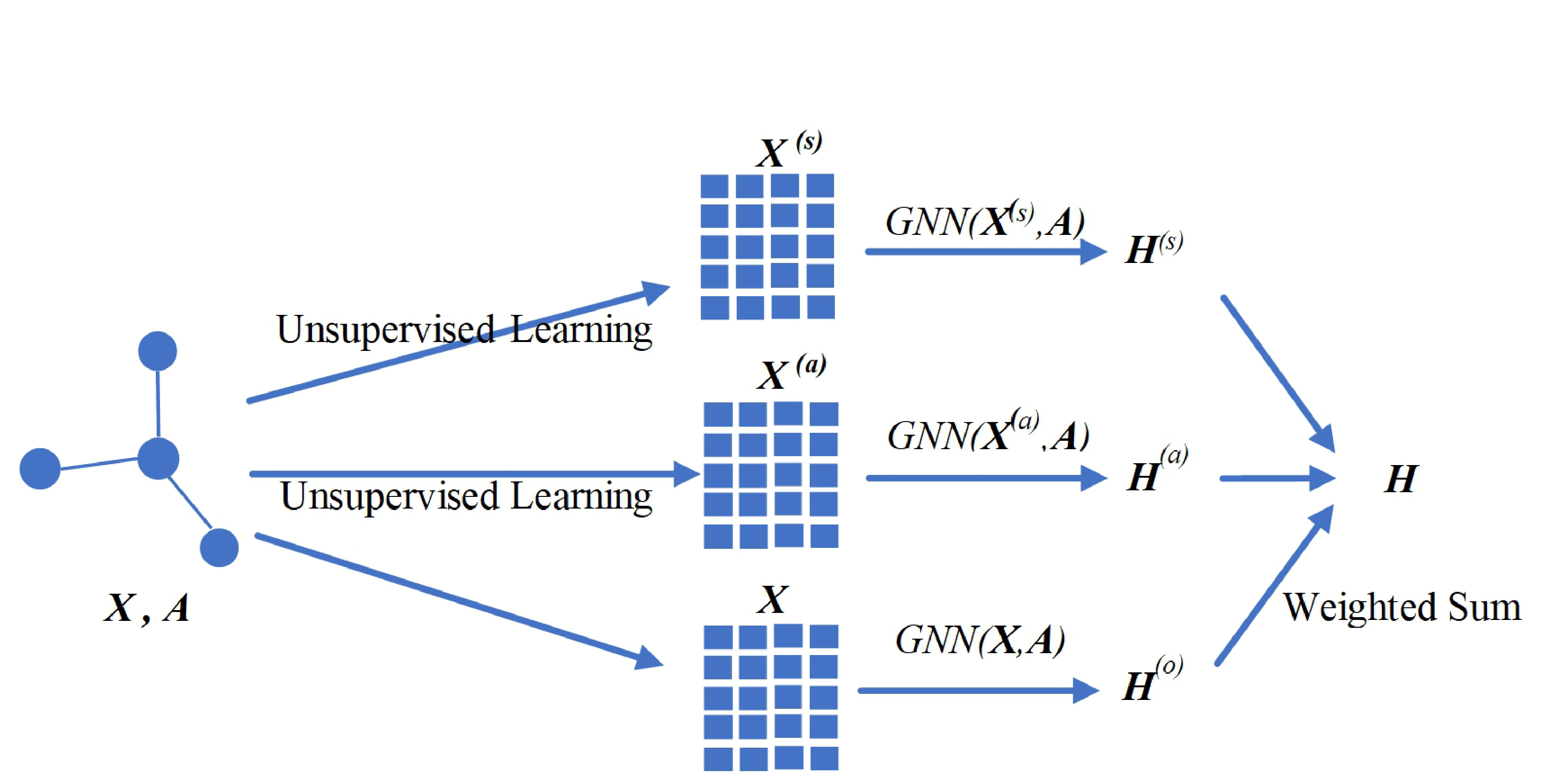}}
\caption{An  overview of the G-GNN framework.}
\end{center}
\end{figure*}

\subsection{Unsupervised Learning of Global Features}
Herein, we propose to learn the unsupervised features of graphs based on random walk strategy. Each node can utilize information within $k$ steps of random walk, where $k$ is often set to $10$. Small world phenomenon suggests that the average distance between nodes will grow logarithmically with the size of the graph nodes~\cite{albert1999internet}, and the undirected average distance of a very large Web graph is only $6.83$ steps~\cite{broder2000graph}. Hence, $10$ steps of random walk can already capture the \emph{global} information of the graphs.
\subsubsection{Global Structure Features}
Similar to Deepwalk~\cite{perozzi2014deepwalk}, the structure features are learned by first sampling the context-source node pairs, and then maximizing their co-occurrence probability. Note that graph attributes $\boldsymbol X$ are not used here. We apply random walks to the graph $G$ to obtain short truncated node sequences. These sequences contain the structure regularity of the original graph. In the node sequences, the neighborhood nodes within certain distance to the source node $v$ are considered as $v$'s context nodes, which are denoted as $N(v) \subset V$.

To maximize the likelihood of the observed source-context pairs, we try to minimize the following objective function:

$$
    \mathop{\prod}_{v \in V}\mathop{\prod}_{u \in N(v)} \frac{ {\rm e}^{\boldsymbol X^{(s)}_v \cdot \boldsymbol X^{(s)}_u}}{\sum\limits_{k \in V} {\rm e}^{\boldsymbol X^{(s)}_v \cdot \boldsymbol X^{(s)}_k}}.
$$
$\boldsymbol X^{(s)} \in \mathbb{R}^{n \times d_s}$ denotes the global structure feature matrix and $d_s$ denotes the dimension of the feature vectors. $\boldsymbol X^{(s)}_v$, $\boldsymbol X^{(s)}_u$ and $\boldsymbol X^{(s)}_k$  denote the global structure feature vectors of $v$, $u$ and $k$ respectively.  The calculation of the denominator is computational expensive since it is required to traverse the entire node set. We approximate it using negative sampling~\cite{Mikolov2013Distributed}.

\subsubsection{Global Attribute Features}
The global attribute features are obtained by maximizing the likelihood of the context attributes. The underlying idea is that if the context attributes can be recovered from the source node, the relationship has already been preserved by the learning model.

For each sampled context node $u \in N(v)$, some attributes of $u$ are sampled as the context attributes of $v$. In this paper, we sample one attribute for one context node. Let $CA(v)$ be the sampled context attributes of $v$, and $T$ be the set of all attributes and $|T|$ is the total number of attributes. We try to minimize the following objective.

$$
    \mathop{\prod}_{v \in V}\mathop{\prod}_{t \in CA(v)} \frac{ {\rm e}^{\boldsymbol X^{(a)}_v \cdot \boldsymbol S_t}}{\sum\limits_{k \in T}{\rm e}^{\boldsymbol X^{(a)}_v \cdot \boldsymbol S_k}}.
$$
where $\boldsymbol X^{(a)} \in  \mathbb{R}^{n \times d_a}$ denotes the global attribute feature matrix, $\boldsymbol S \in \mathbb{R}^{d_a \times |T|}$ denotes the parameters to predict the attributes and $d_a$ denotes the dimension of the attribute feature vectors. $\boldsymbol X^{(a)}_v$ denotes the global attribute feature vector of $v$. $\boldsymbol S_k$ and $\boldsymbol S_t$ denote the corresponding parameter vectors of $k$ and $t$ in $\boldsymbol S$.

Zhu et al.~\cite{zhu2019pcane} proposed an unsupervised graph learning method that utilizes the context attributes. The node representations are learned by jointly optimizing two objective functions that preserve the neighborhood nodes and attributes. The mainly difference of our work is that we learn two feature vectors for each node separately, which provide richer information for the following learning algorithm.

\subsection{Parallel Graph Neural Networks}
As is shown in Fig. 1, we propose a parallel model with kernels of GNN to learn from these input matrices of $\boldsymbol X^{(s)}, \boldsymbol X^{(a)}$ and $\boldsymbol X$. The learning is semi-supervised.
\subsubsection{Learning from the Heterogeneous Features}
The motivation of applying multiple parallel GNN kernels to these feature matrices is as follows. First, the features are quite heterogeneous, especially when some of them are learned via pre-training. The parallel kernels can learn different aspects from these features respectively. Second, the three feature matrices are highly correlated. For example, $\boldsymbol X^{(a)}$ is obtained partly based on $\boldsymbol X$. $\boldsymbol X^{(s)}$ and $\boldsymbol X^{(a)}$ are sampled based on the identical random walk method. It is difficult to learn the complex relationships among them. The parallel setting allows to learn from these features separately, which will make the optimization easier.  Indeed, the parallel schema is successful in some previous papers, such as multi-head attention~\cite{velivckovic2017graph,vaswani2017attention} and N-GCN~\cite{abu2018n}.

First, because the amplitude at each dimension of the pre-trained $\boldsymbol H^{(s)}$ and $\boldsymbol H^{(a)}$ often varies a lot, it is better to make a normalization. For each row $\boldsymbol h$ in $\boldsymbol H^{(s)}$ or $\boldsymbol H^{(a)}$, we make the following transformation, where $mean$ denotes the average function and $std$ denotes the standard derivation function.

$$
    \boldsymbol h = \frac{\boldsymbol h - mean(\boldsymbol h)}{std(\boldsymbol h)}.
$$
Then, several kernels of GNN are proposed to learn from the three feature matrices, as in (4)-(6).

\begin{gather}
    \boldsymbol H^{(s)} = {GNN}(\boldsymbol X^{(s)}, \boldsymbol A).\\
    \boldsymbol H^{(a)} = {GNN}(\boldsymbol X^{(a)}, \boldsymbol A).\\
    \boldsymbol H^{(o)} = {GNN}(\boldsymbol X, \boldsymbol A).
\end{gather}
$GNN(*)$ is the learning kernel of the G-GNN framework. Common GNN-based models that fulfill (3) can be used as a kernel, such as GCN~\cite{kipf2016semi}, Graphsage~\cite{hamilton2017inductive}, APPNP~\cite{klicpera2018predict} and so on. Hence, G-GNN can easily benefit from the strong learning capacities of these kernels.

\subsubsection{Combining the Hidden States}
A simple way to obtain the final hidden state matrix is to linearly combine the three obtained hidden state matrices, where $\alpha$ and $\beta$ are coefficients between 0 and 1.
\begin{equation}
\boldsymbol H = \alpha \boldsymbol H^{(s)} + \beta \boldsymbol H^{(a)} + \boldsymbol H^{(o)}
\end{equation}
Then a softmax function is applied to $\boldsymbol H$, as in (2), to get the prediction probability matrix $\boldsymbol Z$, where $Z_{ij}$ denotes the probability that node $v_i$'s label is the class $j$.
The coefficients of $\alpha$ and $\beta$ are used to turn down the effect of the pre-trained features, which are essential for optimization.  ~\cite{bengio2007greedy} suggests the pre-training is useful for initializing the network in a region of the parameter space where optimization is easier. In training, easy samples often contribute more to the loss and dominate the gradient updating~\cite{lin2017focal}. Similarly, we find the easy-trained components of ${GNN}(\boldsymbol  X^{(s)})$ and ${GNN}(\boldsymbol  X^{(a)})$ also dominate the learning procedure. If no weight strategy is used, ${GNN}(\boldsymbol  X)$ merely contributes to the results and hence the performance is far from promising.

\subsubsection{Training}
We minimize the cross-entropy loss function between $\boldsymbol Z$ and the ground-truth labels $\boldsymbol Y$ to train the model~\cite{abu2018n}.

$$
    \min {diag}(\boldsymbol V_{\rm  train})[\boldsymbol Y \circ  \log \boldsymbol Z]
$$
where $\circ$ denotes Hadamard product, and ${diag}(\boldsymbol V_{\rm  train})$ denotes a diagonal matrix, with entry at $(i,i)$ set to 1 if $v_i \in \boldsymbol V_{\rm  train}$ and 0 otherwise.

\subsection{Learning on Plain Graphs}
Plain graphs contain no attributes $\boldsymbol X$. It does not affect the obtaining of the global structure feature matrix $\boldsymbol X^{(s)}$. Then the final hidden states is:
$$
    \boldsymbol H = \boldsymbol H^{(s)} = {GNN}(\boldsymbol X^{(s)}, \boldsymbol A).
$$

Hence, learning on plain graphs also follows the pretraining-and-learning schema, where some components that depend on the graph attributes are removed.

\subsection{Discussion}
In the paper, we choose a parallel framework to learn from these heterogeneous attributes. But are there any alternative technique choices?

The pretrained vectors can be integrated with the classification model in different stages, e.g. early/middle/late fusion.
\begin{itemize}
    \item Early fusion:  Early fusion is quite simple, where the pretrained and raw attributes are combined in the input layer. For example, we can concatenate the attributes and then feed them to a GNN model.
    \item Middle fusion: Middle fusion is a bit more complicated. It is required to design specifical propagation and aggregation functions of GNN to integrate the features in each layer.
    \item Late fusion: The late fusion is the current version of parallel framework. The attributes are fed to different GNNs and the outputs are combined in the final layer.
\end{itemize}

In practice, early fusion can already improve the precision, but the improvement is not so big as late fusion. See table 3 in section 4.2.2 for details. Middle fusion requires to modify the inner propagation or aggregation functions of the specifical GNN. Hence, the framework will lose generality to apply to other GNNs. Therefore, we choose late fusion as our learning framework, which not only gives the best performance, but also can improve the global information utilization ability of general GNNs.

\section{Experiments}

In this section, we conduct experiments to answer the following research questions\footnote{The data, code and pre-trained vectors to reproduce our results are released on https://github.com/zhudanhao/G-GNN}:
\begin{itemize}
    \item \textbf{Q1}: How does G-GNN perform in comparison with state-of-the-art GNN kernels on attributed graphs?
    \item \textbf{Q2}: Are all the designed components in G-GNN helpful for achieving stronger learning ability?
    \item \textbf{Q3}: How does G-GNN perform in comparison with state-of-the-art learning methods on plain graphs?

\end{itemize}

\subsection{Experiments on Attributed Graphs (Q1)}
Herein, we will answer  \textbf{Q1} by comparing G-GNN with different GNN kernels. Note that the codes of all the GNN methods are based on the implementation released by  DGL~\footnote{https://github.com/dmlc/dgl/tree/master/examples/pytorch}. All results are the average values of 10 experiments with different random initialization seeds.
\subsubsection{Datasets and Baselines}

\vspace{2mm}
\tabcolsep 4pt
\renewcommand\arraystretch{1.3}
\vspace{2mm}
\begin{table*}
\caption{ The statistics of the datasets}
\centering
\footnotesize{
\begin{tabular}{cccccccc}\hline\hline\hline
\hline
 Dataset  &\#Nodes &\#Edges &\#Attributes &\#Classes &\#Training nodes&\#Valid nodes&\#Test nodes\\
\hline
Cora&2708&5429&1433&7&140&500&1000\\
Citeseer&3327&4732&3703&6&120&500&1000\\
Pubmed&19717&44338&500&3&60&500&1000\\
CoraFull&19793&130622&8710&70&1792&1792&15773\\

\hline
\end{tabular}
}
\\Note: \# denotes the number of something, e.g. \#Nodes is the number of nodes.

\end{table*}

The statistics of the datasets used in this study are shown in Table 1. The three standard attributed graph benchmark datasets of Cora, Citeseer and Pubmed~\cite{sen2008collective} are widely used in various GNN studies~\cite{velivckovic2017graph,yang2016revisiting}. In the citation graphs, nodes denote papers and links denote undirected citations. Node attributes are the extracted elements of bag-of-words representation of the documents. The class label is the research area of each paper. CoraFull is an extended version of Cora~\cite{bojchevski2017deep}. Following ~\cite{bojchevski2017deep}, we randomly split the train/valid/test dataset with 1:1:8.
\setcounter{table}{1}
\begin{table*}
\label{tab:5.3}
\centering
\caption{Classification Precision (\%) on Attributed Graphs}
{\footnotesize{
\begin{tabular}{ccccccccc}
\hline

\multirow{2}{*}{Method}&\multicolumn{2}{c}{Cora}&\multicolumn{2}{c}{Citeseer}&\multicolumn{2}{c}{Pubmed}&\multicolumn{2}{c}{CoraFull}\\
&Precision&Range&Precision&Range&Precision&Range&Precision&Range\\
\hline
GCN~\cite{kipf2016semi}& 81.47& $\pm 2.60$&71.01&$\pm 1.30$&79.10&$\pm 1.10$&59.44&$\pm 1.02$\\
G-GCN&83.71&$\pm 1.80$&71.27&$\pm 1.40$&80.88&$\pm 0.95$&61.20&$\pm 0.37$\\
\hline
Graphsage~\cite{hamilton2017inductive}&82.70&$\pm 1.65$&69.87&$\pm 1.30$&78.56&$\pm 0.65$&59.21&$\pm0.59$\\
G-Graphsage&83.84&$\pm 1.10$&70.20&$\pm 1.15$&78.89&$\pm 1.85$&60.99&$\pm 0.71$\\
\hline
GAT~\cite{velivckovic2017graph}& 82.39&$\pm 0.97$&67.52 &$\pm 1.74$ &77.34&$\pm 0.92$&62.65&$\pm0.44$\\
G-GAT& 82.53&$\pm 1.12$&67.95&$\pm 1.38$ &76.88&$\pm 0.74$&\textbf{63.42}&$\pm 0.38$\\
\hline
APPNP~\cite{klicpera2018predict}&83.91&$\pm 1.45$&71.93&$\pm 1.55$&79.68&$\pm 0.65$&61.71&$\pm0.7$\\
G-APPNP&\textbf{84.31}&$\pm 1.30$&\textbf{72.00}&$\pm 1.45$&\textbf{80.95}&$\pm 0.80$&63.19&$\pm 0.84$\\
\hline
\end{tabular}
}}
\end{table*}

The following baselines are compared in this paper.
\begin{itemize}
    \item Graph convolutional network (GCN)~\cite{kipf2016semi}: It is a simple type of GNN introduced in details in section 2. We use dropout technique to avoid overfitting~\cite{Srivastava2014Dropout}, where the probability is $0.5$. We set the number of training epoches to 300, the number of layers to 2, and the dimension of hidden states to 16. The self-loops are used.
    \item Graphsage~\cite{hamilton2017inductive}: It is a general framework by sampling and aggregating features from a node's local neighborhood.  We use the mean aggregate. We set the dropout rate to 0.5, the number of training epoches to 200, the number of layers to 2, and the dimension of hidden states to 16.
    \item APPNP~\cite{klicpera2018predict}: APPNP is designed with a new propagation procedure based on personalized PageRank, and hence can also model the long-distance information to a source node. We set the dropout rate to 0.5, the number of training epoches to 300, the number of propagation steps to 10, the teleport probability to 0.1 and the dimension of hidden states to 64.
    \item Graph attention network (GAT)~\cite{velivckovic2017graph}: GAT is designed with the multi-head attention techniques~\cite{vaswani2017attention} to the aggregation method, and hence can attribute different weights  to different neighbor nodes. We set the dropout rate to 0.6, the number of heads to 8, the number of hidden states to 8, the number of layers to 2 and the number of training epoches to 200.
\end{itemize}
All models are optimized with Adam~\cite{Kingma2014Adam} where the initial learning rate is 0.01 and the weight decay is 0.0005 per epoch.

\subsubsection{Training Details}
In the unsupervised learning of global features, we conduct 10 iterations of random walk start from each node. The walk length is 100. For each source node, the nearby nodes within 10 steps are considered as the neighborhood nodes. The dimensions of both the global structure and attribute vectors are 8. The number of negative sampling is 64.

In the semi-supervised learning, the three baseline models are used as kernels of G-GNN, and the corresponding models are named as G-GCN, G-Graphsage and G-APPNP. The parameters are exactly the same as those in the baseline methods. We search $\alpha$ and $\beta$ between 0.001 to 0.05. The test results are reported when the best valid results are obtained.

\begin{figure*}[!htp]
\label{fig:2}
\begin{center}
\subfigure[]{
\begin{minipage}[t]{0.3\linewidth}
\centering
\includegraphics[width=150pt]{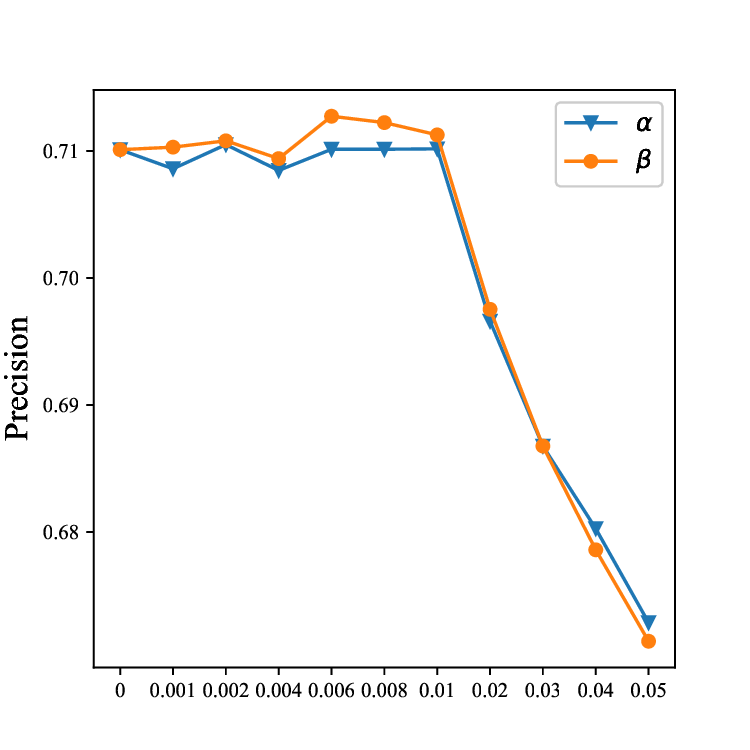}
\end{minipage}%
}%
\subfigure[]{
\begin{minipage}[t]{0.3\linewidth}
\centering
\includegraphics[width=150pt]{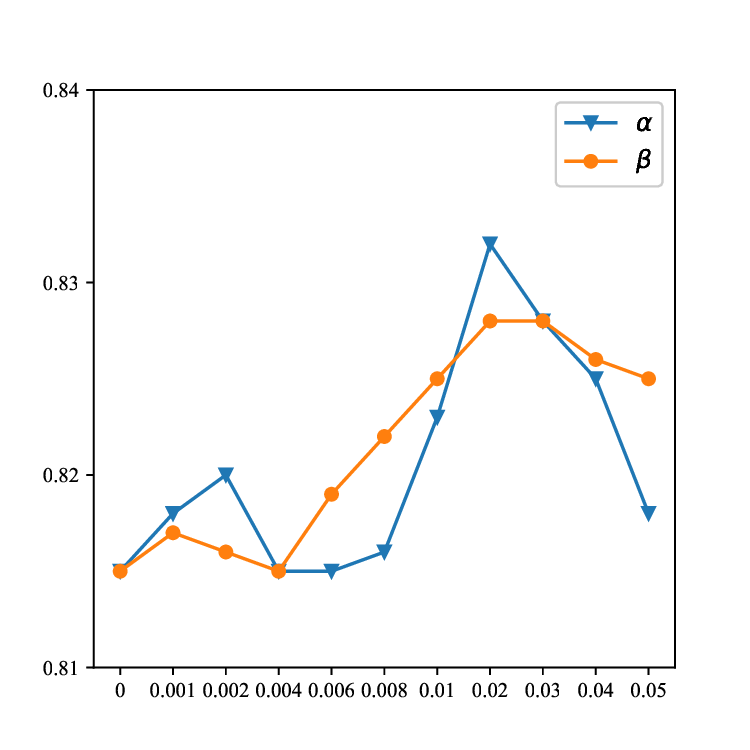}
\end{minipage}%
}%
\subfigure[]{
\begin{minipage}[t]{0.3\linewidth}
\centering
\includegraphics[width=150pt]{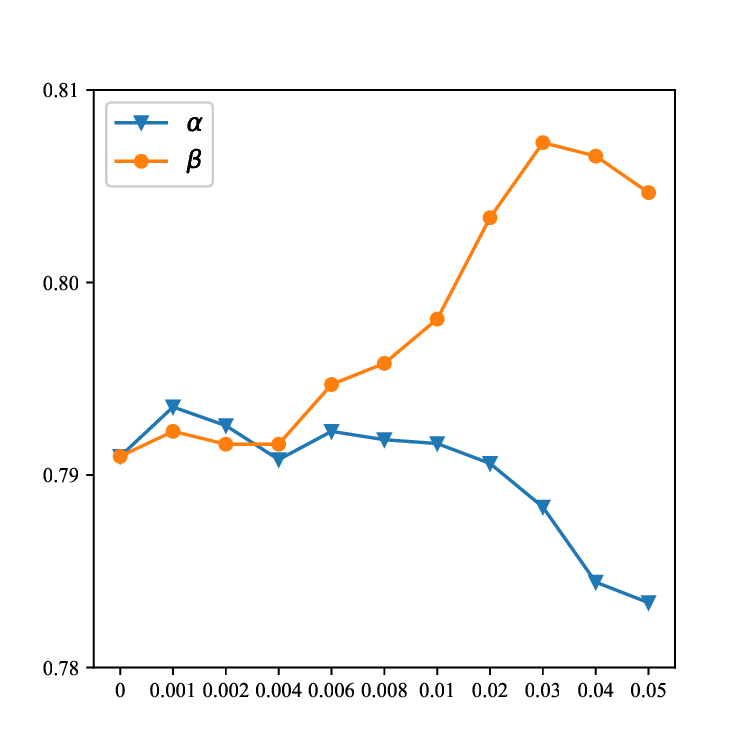}
\end{minipage}%
}%
\caption{Precision w.r.t. $\alpha$ and $\beta$ on different datasets. (a) Citeseer. (b) Cora. (c) Pubmed.}
\end{center}
\end{figure*}

\subsubsection{Results}
The results are shown in Table 2. From the table, it is found that all baseline kernels with global information achieve substantial gains on the classification task. For example, G-GCN outperforms GCN with 2.24\%, 0.26\%, 1.79\% and 1.76\% of precision on the four datasets respectively. The results demonstrate that the learning framework of G-GNN can effectively and consistently enhance the learning ability of the corresponding GNN kernels.

APPNP is also designed for enlarging the reception field and can utilize global information. APPNP outperforms the other baseline models. Although the improvement is not so large as those in G-GCN, G-APPNP still significantly outperforms its kernel of APPNP with 0.40\%, 0.07\%, 1.27\% and 1.48\% on the four datasets respectively. The result shows that even for a propagation method which can powerfully utilize global information, our learning schema of G-GNN can still bring considerable precision gains. We believe the advantage comes from the pretraining-and-learning schema, since our global information is obtained via pre-training and no longer suffers from the limitation brought by weak supervision.

G-APPNP achieves the best results on Cora, Citeseer and Cora. Note that its precisions on Cora (84.31\%) and Pubmed (80.95\%) are the new state-of-the-art results. To the best of our knowledge, the previous best results are GraphNAS (84.20\%)~\cite{gao2019graphnas} on Cora, and MixHop (80.80\%) ~\cite{abu2019mixhop} on Pubmed. On the large dataset of CoraFull, G-GAT outperforms all the other methods with precision of 63.42\%.

In all, the results validate the effectiveness of the pretraining-and-learning schema, which can significantly improve the global information preserving ability of GNN based methods.

\subsection{Properties Analysis (Q2)}
The parameter sensitivity and ablation analysis are given. Herein, we mainly use GCN as the kernel and the setting is attributed graph learning.

\subsubsection{Parameter Sensitivity}
Fig. 2 shows the precision w.r.t. $\alpha$ and $\beta$. Generally, different datasets require different $\alpha$ and $\beta$ to achieve the best precision, and $\alpha$ and $\beta$ are often around 0.01. The precision will decrease quickly if we continue to increase the two parameters. The result shows that it is very necessary to introduce the two hyper-parameters to turn down the impact of the pre-trained features. In fact, the component of ${GNN}(\boldsymbol X, \boldsymbol A)$ will contribute almost nothing without the weight method.

Fig. 3 shows the precision w.r.t. the dimension of the global features. The highest precision is achieved when the dimension is around 8 to 16.

\vspace{2mm}

\begin{center}
{\includegraphics[width=0.7\linewidth]{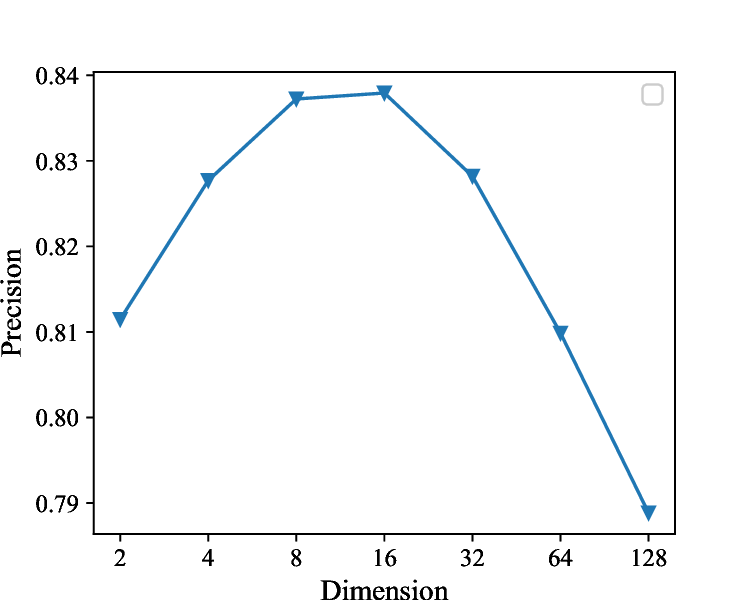}}
\parbox[c]{8.3cm}{\footnotesize{Fig.3.~}  Precision w.r.t. Dimension of global features on Cora. }
\end{center}
\subsubsection{Ablation Analysis}
First, the effectiveness of parallel learning method is investigated. Note that the simplest way to utilize all three feature matrices $\boldsymbol X^{(s)}$, $\boldsymbol X^{(a)}$ and $\boldsymbol X$ is to concatenate them first, and then feed the concatenated feature matrix to a single GNN kernel, so called early fusion in subsection 3.5. Table 3 compares the results of GCN, early fusion and G-GCN. We can find that early fusion has already outperforms GCN, which demonstrates that the pretraining-and-learning schema of G-GCN can well utilize the global information. G-GCN makes further improvement than the method of simple concatenation, which validates the effectiveness of parallel learning method.

\begin{center}
{\footnotesize{\bf Table 3.} The classification precision (\%) on different feature fusion settings }\\
\vspace{2mm}
{\footnotesize

\begin{tabular}{cccc}
\hline
Method&Cora&Citeseer&Pubmed \\
\hline
GCN&81.47&71.01&79.10\\
Early Fusion&83.39&70.78&80.64\\
Late Fusion (G-GCN)&83.71&71.27&80.88\\
\hline
\end{tabular}

}
\end{center}
Second, we investigate the effect of the global features, as shown in Table 4. The results show that both pre-trained feature matrices can help to increase the model precision. The global structure features are more helpful on Cora, but less effective than global attribute features on Pubmed and Citeseer. On the datasets of Cora and Pubmed, the highest precisions are obtained when all the features are used. However, in Citeseer, the global structure features cannot help to increase the performance if the global attribute features are used. In all, both pre-trained features can contribute to improve the results of our proposed model. However, the amount of improvement depends on specific datasets.
\setcounter{table}{3}
\begin{table*}
\label{tab:4}
\centering
\caption{The classification precision (\%) when using different feature matrices }
{\footnotesize
\begin{tabular}{cccc}
\hline

Feature Matrices&Cora&Citeseer&Pubmed \\
\hline
$\boldsymbol X$&81.47&71.01&79.10\\
$\boldsymbol X+\boldsymbol X^{(s)}$&83.20&71.05&79.35\\
$\boldsymbol X+\boldsymbol X^{(a)}$&82.77&71.27&80.73\\
$\boldsymbol X+\boldsymbol X^{(s)}+\boldsymbol X^{(a)}$&83.71&71.27&80.88\\
\hline
\end{tabular}
\\
Note: $\boldsymbol X^{(s)}$, $\boldsymbol X^{(a)}$, $\boldsymbol X$ are the global structure, the global attribute, and the raw graph features respectively. The model is G-GCN. Each line gives the results of G-GCN with the corresponding feature matrices. Hence, the model in the first line is equivalent to GCN, and the one of the last line is corresponding to the full G-GCN model.
}
\end{table*}

\subsection{Experiments on Plain Graphs (Q3)}

\begin{table*}
\label{tab:5.2}
\centering
\caption{Classification precision (\%) on plain graphs}
{\footnotesize
\begin{tabular}{ccccccccccc}
\hline
   \multirow{2}{*}{Dataset}&\multirow{2}{*}{Method}& \multicolumn{9}{c}{Training percent}\\ &&10\% &20\% &30\% &40\% &50\% &60\% &70\% &80\% &90\% \\
\hline
&ICA~\cite{sen2008collective}&44.96&47.21&47.40&48.23&48.82&50.18&52.43&58.05&66.04\\
Cora&Planetoid-G~\cite{yang2016revisiting}&74.45&79.16&80.72&82.63&84.59&83.34&84.23&84.67&85.10 \\

&PNE~\cite{chen2017pne}&\textbf{77.58}&\textbf{81.22}&\textbf{82.94}&84.54&84.73&85.55&86.15&86.39&87.76 \\

&MMDW~\cite{Tu2016Max}&74.94&80.83&82.83&83.68&84.71&85.51&87.01&\textbf{87.27}&88.19\\

&G-GCN (Plain)&76.88 &80.5&82.65&\textbf{85.06}&\textbf{85.57}&\textbf{86.23}&\textbf{87.67}&87.05&\textbf{89.16}  \\
\hline

\multirow{3}{*}{Citeseer} &ICA~\cite{sen2008collective}&33.54&34.95&37.59&38.36&40.92&42.79&46.23&48.96&57.05 \\
&Planetoid-G~\cite{yang2016revisiting}&54.10&57.92&58.15&62.31&64.22&68.11&70.16&70.34&72.12 \\
&PNE~\cite{chen2017pne}&54.79&60.87&\textbf{64.67}&\textbf{66.95}&68.59&70.00&72.06&73.41&74.76\\
&MMDW~\cite{Tu2016Max}&\textbf{55.60}&\textbf{60.97}&63.18&65.08&66.93&69.52&70.47&70.87&70.95\\
&G-GCN (Plain)&54.24&60.31&64.16&66.41&\textbf{69.36}&\textbf{70.77}&\textbf{72.12}&\textbf{74.41}&\textbf{75.89} \\
\hline

\end{tabular}
\\
Note: The results of PNE~\cite{chen2017pne} and MMDW~\cite{Tu2016Max} are cited from their original papers.
}
\end{table*}

We will answer \textbf{Q3} by comparing G-GNN with other plain graph learning methods.
\subsubsection{Experiment Setup}
We conduct the task of semi-supervised classification on the two datasets of Cora and Citeseer. The results of the baseline methods are cited from their original papers. Pubmed is excluded from comparison since it is not used in the baseline papers.

The training data is the entire plain graph ($\textbf{X}$ is excluded), and part of the node labels. We use 0.1, 0.2 ... 0.9 of node labels to train the model respectively, and report the classification accuracy on the rest of data. All results are the average values of 10 experiments with different random split.

Our proposed model is G-GCN (plain), where the GCN is used as the kernel. In the unsupervised training, the dimension of global structure vectors is 32. In the semi-supervised learning, the dimension of the hidden states is 256. No dropout is used.  The rest of parameters are the same as those in 4.1.2.

Four semi-supervised learning methods on plain graphs are used as baselines.
\begin{itemize}
    \item Iterative Classification Algorithm (ICA)~\cite{sen2008collective}: ICA iteratively propagates observed labels to nearby unlabeled nodes, until the assignments to the labels stabilize. Here we use k-nearest neighbor as the classification algorithm.
    \item Planetoid-G~\cite{yang2016revisiting}: The method trains an embedding for each instance to jointly predict the class label and the neighborhood context in the graph.
    \item MMDW~\cite{Tu2016Max}: The method jointly optimizes the max-margin classifier and the aimed social representation learning model.
    \item PNE~\cite{chen2017pne}: The method embeds nodes and labels into the same latent space.
\end{itemize}

Some other unsupervised or semi-supervised methods, such as LP~\cite{zhu2003semi},  Deepwalk~\cite{perozzi2014deepwalk}, Line~\cite{tang2015line} and LSHM~\cite{Jacob2014Learning} are excluded from comparison since the baseline methods have demonstrated that they are outperformed by the baseline methods~\cite{yang2016revisiting,Tu2016Max,chen2017pne}.

\subsubsection{Results}
The results are shown in Table 5. G-GCN (plain) achieves the highest precision on more than half of the total data points (10 out of 18). Specially, the advantage is more obvious when the training ratio is arising. When the training ratio is small, e.g. less than 30\%, the categories with less instances may provide very few training instances, which makes GNN difficult to pass message from these nodes. We believe this is the reason why G-GCN is less powerful when training ratio is small.

In all, the results show that the learning framework of G-GNN can be successfully applied to plain graphs, and achieve similar or better results than state-of-the-art methods.

\section{Related Work}
There are a lot of efforts in recent literature to develop neural network learning algorithms on graphs. The most prominent one may be Graph convolution networks (GCN)~\cite{kipf2016semi}. GCN is based on the first-order approximation of spectral graph convolutions. Graphsage~\cite{hamilton2017inductive} is new neighborhood aggregation algorithms by concatenating the node's features in addition to pooled neighborhood information. Graph attention model (GAT)~\cite{vaswani2017attention} was proposed to assign different neighborhoods with different weights based on multi-head attention mechanism. In FastGCN~\cite{chen2018fastgcn},  graph convolutions is interpreted as integral transforms of embedding functions under probability measures, which has faster training speed and comparable precision. P-PGNNs~\cite{you2019position} can capture nodes' position within graph structure, which first sample sets of anchor nodes and then learn a non-linear distance-weighted aggregation scheme over the anchor-sets.

Several related studies tried to expand the reception field of GNN and increase the neighborhood available at each node. PPNP/APPNP~\cite{klicpera2018predict} improves the message passing algorithms based on personalized PageRank. ~\cite{xu2018representation} proposes jumping knowledge networks that can that flexibly leverage different neighborhood ranges for each node. N-GCN~\cite{abu2018n} trains multiple instances of GCNs over node pairs discovered at different distances in random walks, and learns a combination of these instance outputs. However, because the semi-supervised settings lack enough training, these methods have to control the model complexity carefully, which limits the learning ability in exploring the global information. For example, our experiments have shown that G-GNN with kernel of APPNP can still achieve promising improvement.

Some studies also try to introduce unsupervised learning in GNNs to alleviate the insufficient supervision problem. ~\cite{Tran-LoNGAE:2018} proposes an auto-encoder architecture that learns a joint representation of both local graph structure and available node features for the multi-task learning of link prediction and node classification. GraphNAS\cite{gao2019graphnas} first generates variable-length strings that describe the architectures of graph neural networks, and then maximizes the expected accuracy of the generated architectures on a validation data based on reinforcement learning. However, these methods do not consider to utilize global information of the graphs. The main difference of our work is that we use unsupervised learning to capture the global information.

\section{Conclusion}
In the paper, we proposed a novel framework named G-GNN, which is able to conduct semi-supervised learning on both plain and attributed graphs. The proposed framework takes the advantage of both random walk and GNN, which can not only utilize global information but also aggregate local information well. Therefor, the existing GNN models can be used as kernels of G-GNN, to obtain the ability of preserving global information. Extensive experiments show that our framework can improve the learning ability of existing GNN models.

For future work, we plan to test some more complicated methods that combine the hidden states, and study other unsupervised methods that can produce global features more suitable for the learning ability of GNN.

\vspace{5mm}

\noindent\parbox{8.3cm}{\parpic{\includegraphics[width=80px]{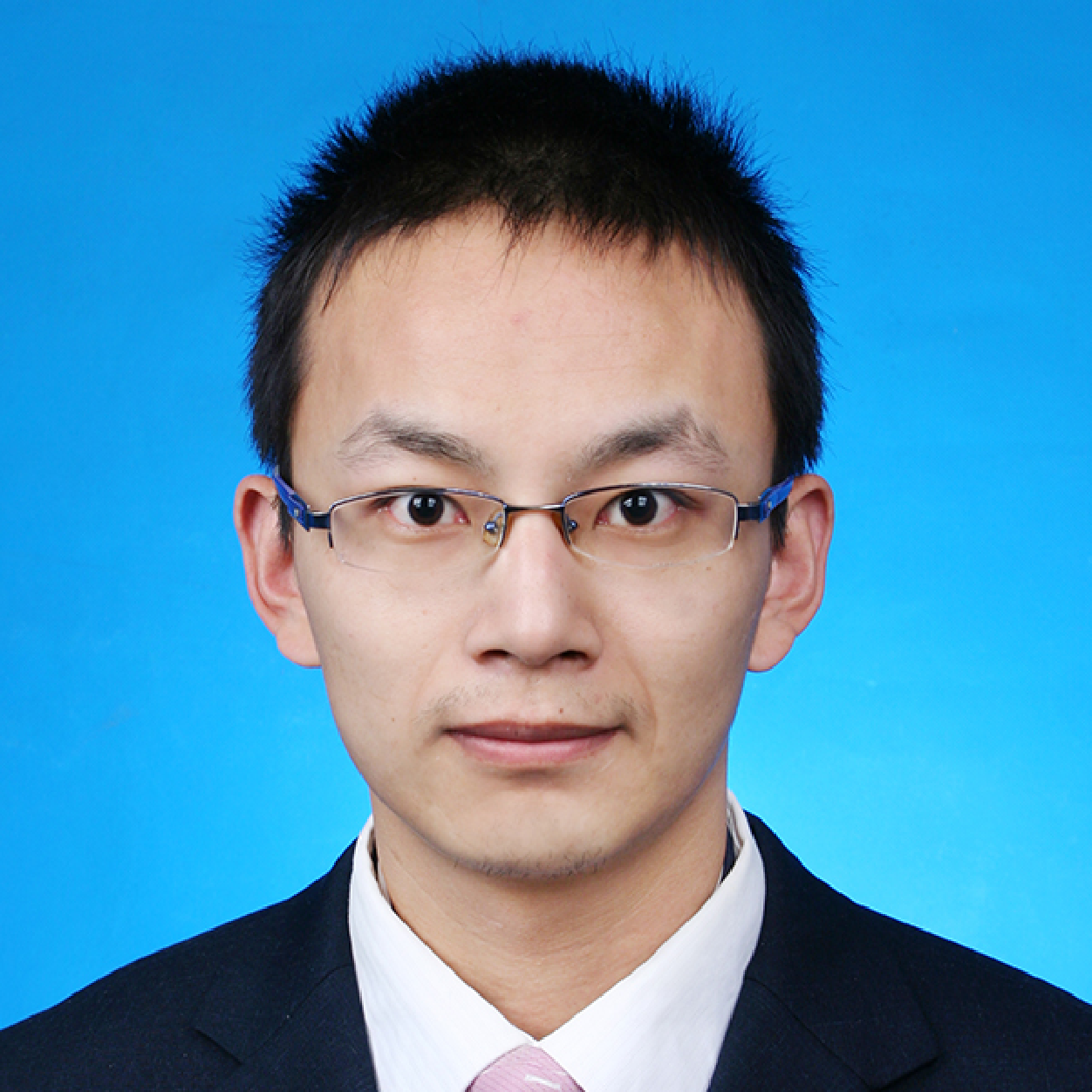}}{\small\quad   \textcolor{blue}{Dan-Hao Zhu received his B.S. degree in Management  from Hohai University, Changzhou, in 2008, M.S. in Management  from Nanjing University, Nanjing, in 2011, and Ph.D.  in computer science in Nanjing University, Nanjing, in 2019. He is currently a librarian in Jiangsu Police Institute. His research interests include network representation, graph learning and natural language processing. }}\\[1mm]}

\noindent\parbox{8.3cm}{\parpic{\includegraphics[width=80px]{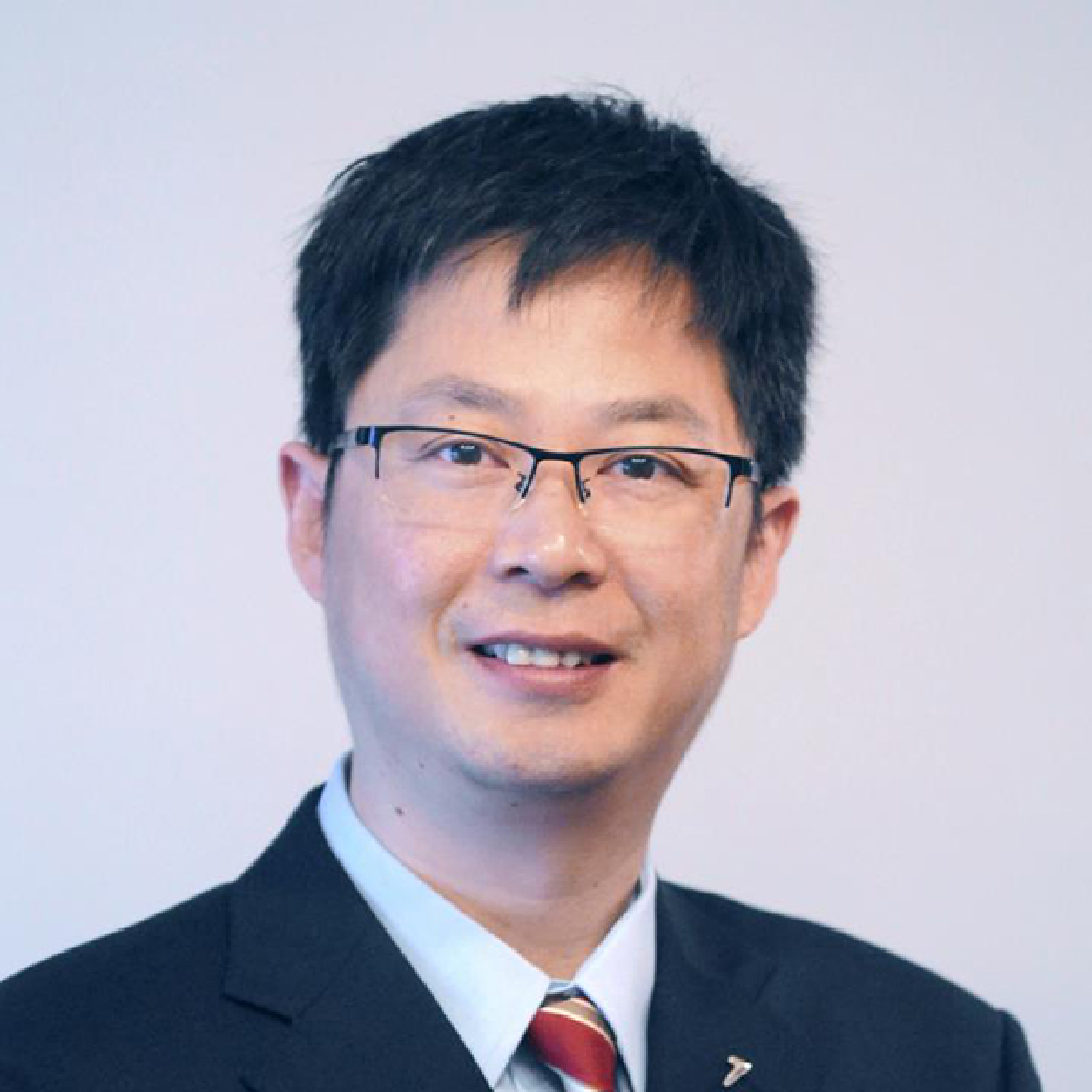}}{\small\quad   \textcolor{blue}{Xin-Yu Dai received the B.Eng. and Ph.D. degrees in computer science, Nanjing University, Nanjing, in 1999 and 2005, respectively. He has been on leave from Nanjing University from August 2010 to September 2011 to visit EECS Department and Statistics Department at UC Berkeley. He is currently a professor with the School of Artificial Intelligence at Nanjing University. His research interests include natural language processing and knowledge engineering.} }\\[1mm]}

\noindent\parbox{8.3cm}{\parpic{\includegraphics[width=80px]{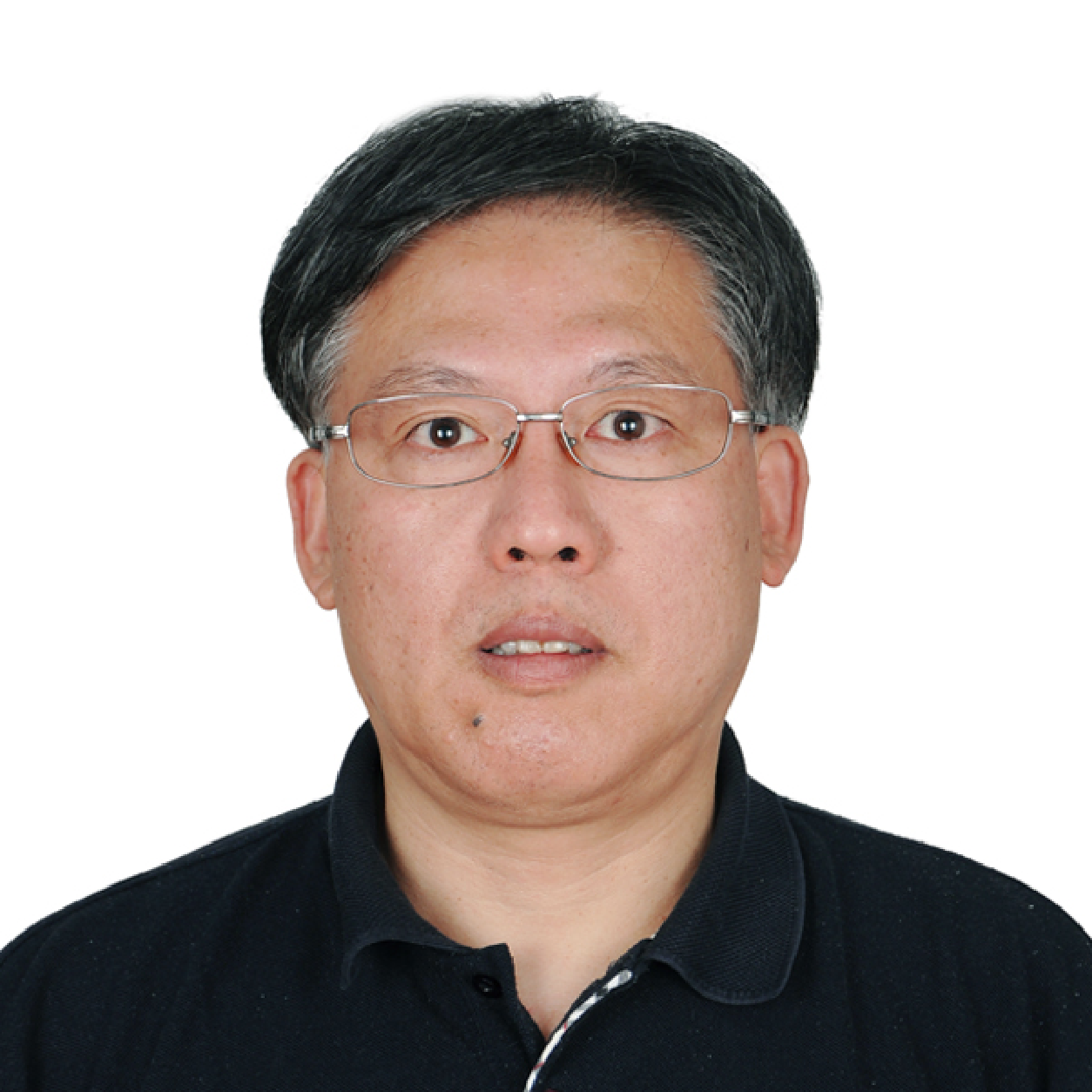}}{\small\quad  \textcolor{blue}{Jia-Jun Chen is a professor of Department of Computer Science and Technology, Nanjing University, and the director of Natural Language Processing Group. Prof. Chen received his Ph.D., M.S. and B.S. in computer science from Nanjing University, Nanjing, in 1985, 1988, 1998 respectively. His research interest is on natural language processing, machine translation,  information extraction, and text classification. } }\\[1mm]}

\label{last-page}
\end{multicols}
\label{last-page}
\end{document}